\pdfoutput=1

\documentclass[11pt]{article}

\usepackage{EMNLP2023}

\usepackage{times}
\usepackage{latexsym}

\usepackage[T1]{fontenc}

\usepackage[utf8]{inputenc}

\usepackage{microtype}

\usepackage{inconsolata}

\usepackage{multirow}
\usepackage{multicol}
\usepackage{microtype}
\usepackage{bbding}
\usepackage{bm}
\usepackage{colortbl}

\usepackage{amsmath}
\usepackage{amssymb}
\usepackage{algorithm}
\usepackage{algorithmic}
\usepackage{amsthm}
\usepackage{cleveref} 
\usepackage{subfig}
\usepackage{supertabular}
\usepackage{longtable}  
\usepackage{tabularray}  
\usepackage[utf8]{inputenc}
\UseTblrLibrary{booktabs}

\usepackage[american]{babel}

\usepackage{color}
\usepackage{graphicx}
\allowdisplaybreaks

\newcommand{\eg}{\textit{e.g.}}
\newcommand{\ie}{\textit{i.e.}}

\usepackage{booktabs}
\usepackage{multirow}
\usepackage{graphicx}

\def \video {\mathbf{v}}
\def \text {\mathbf{t}}
\def \word {w}

\def \feature {\mathbf{r}}
\def \framefeature {\feature^{f}}
\def \videofeature {\feature^{v}}
\def \sentencefeature {\feature^{s}}

\def \framecodebook {\feature^{f}_c}
\def \videocodebook {\feature^{v}_c}
\def \sentencecodebook {\feature^{s}_c}

\def \videocodebooksim {sim^{\video}}

\def \framecodebooksim {sim^{f}}

\def \videotosentencesim {S_{\videofeature-\sentencefeature}}
\def \frametosentencesim {S_{\framefeature-\sentencefeature}}
\def \videotosentencecodebooksim {S_{\videocodebook-\sentencecodebook}}
\def \frametosentencecodebooksim {S_{\framecodebook-\sentencecodebook}}

\def \videotosentenceattention {A_{\videofeature-\sentencefeature}}
\def \frametosentenceattention {A_{\framefeature-\sentencefeature}}

\def \videotosentencecodebookattention {A_{\videocodebook-\sentencecodebook}}
\def \frametosentencecodebookattention {A_{\framecodebook-\sentencecodebook}}

\def \sentencecodebooksim {sim^{\text}}

\def \sim {S}
\newcommand{\ours}{\textsc{S3MA}}

\title{Video-Text Retrieval by Supervised Sparse Multi-Grained Learning}

\author{Yimu Wang\\
  University of Waterloo \\
  \texttt{yimu.wang@uwaterloo.ca} \\\And
  Peng Shi \\
  University of Waterloo \\
  \texttt{peng.shi@uwaterloo.ca} \\}

\begin{document}
\maketitle
\begin{abstract}
While recent progress in video-text retrieval has been advanced by the exploration of better representation learning, in this paper, we present a novel multi-grained sparse learning framework, \ours, to learn an aligned sparse space shared between the video and the text for video-text retrieval. 
The shared sparse space is initialized with a finite number of sparse concepts, each of which refers to a number of words. 
With the text data at hand, we learn and update the shared sparse space in a supervised manner using the proposed similarity and alignment losses. 
Moreover, to enable multi-grained alignment, we incorporate frame representations for better modeling the video modality and calculating fine-grained and coarse-grained similarities. 
Benefiting from the learned shared sparse space and multi-grained similarities, extensive experiments on several video-text retrieval benchmarks demonstrate the superiority of \ours~over existing methods. 
Our code is available at \href{https://github.com/yimuwangcs/Better_Cross_Modal_Retrieval}{link}.
\end{abstract}

\section{Introduction}

\begin{figure}[t!]
\centering
\includegraphics[width=1\columnwidth]{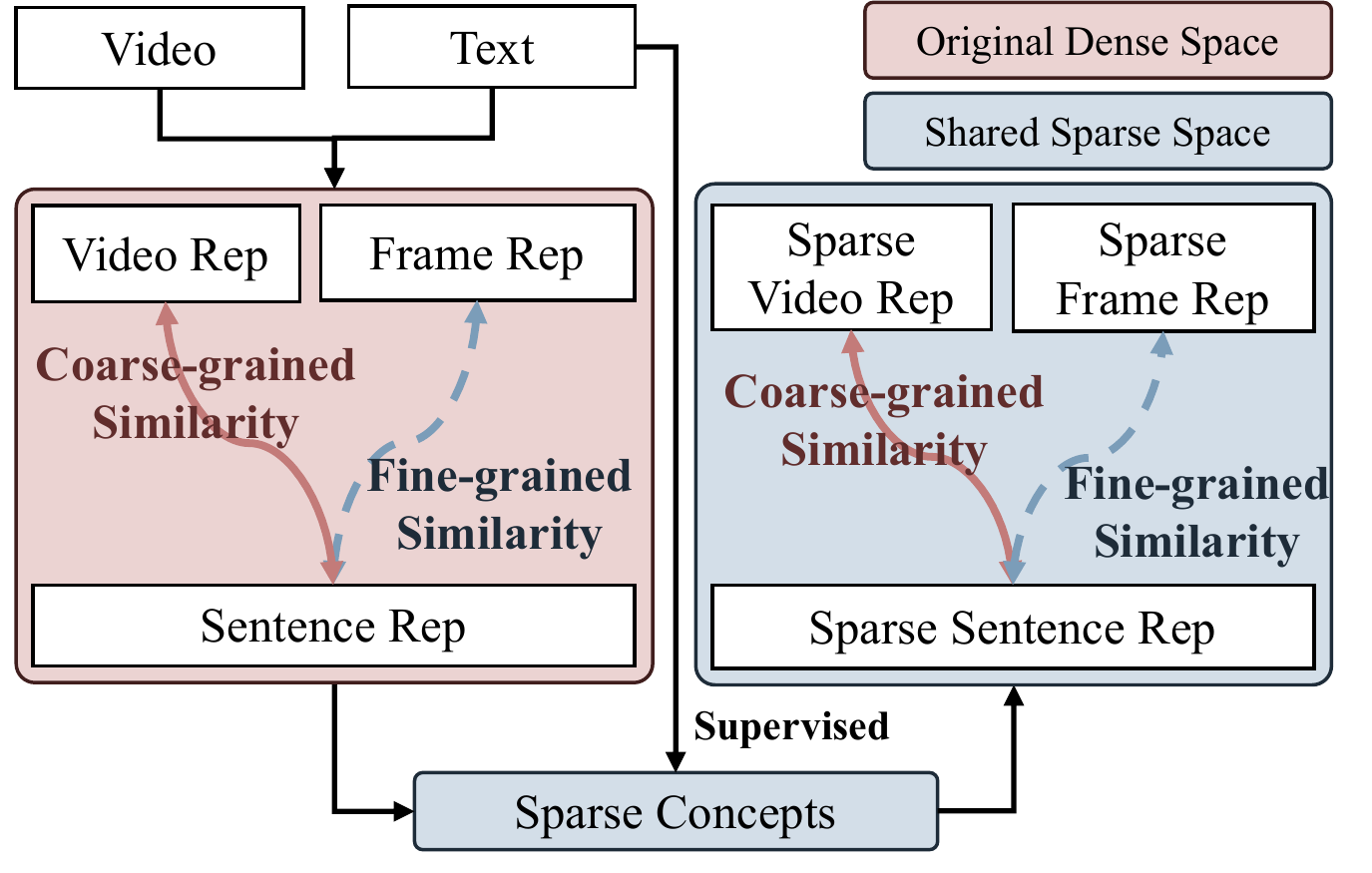}
\caption{
Our proposed \textit{supervised shared sparse multi-grained alignment} framework for video-text retrieval maps sentence, video, and frame representations to a shared sparse space to obtain sparse sentence, video, and frame representations. 
Then, it calculates \textit{coarse- and fine-grained} similarities to fully explore the power of the sparse space, which is learned in a \textit{supervised} fashion. 
``Original Dense Space'' represents the space containing the representations generated from modality-dependent encoders. 
``Shared Sparse Space'' represents the space containing the sparse concepts shared across two modalities. 
``Rep'' refers to representation.
}
\end{figure}
As a fundamental task in visual-language understanding~\cite{9093306,xu-etal-2021-videoclip,park-etal-2022-exposing,miyawaki-etal-2022-scene,fang2023you,fang2023hierarchical,kim2023nice}, video-text retrieval (VTR)~\cite{DBLP:journals/ijon/LuoJZCLDL22,DBLP:journals/corr/abs-2111-05610,DBLP:conf/mm/MaXSYZJ22,liu-etal-2022-cross,10.1145/3477495.3531950,DBLP:conf/cvpr/GortiVMGVGY22,fang2022multi} has attracted interest from academia and industry.
Although recent years have witnessed the rapid development of VTR with the support from powerful pretraining models~\cite{DBLP:journals/ijon/LuoJZCLDL22,DBLP:journals/corr/abs-2111-05610,DBLP:conf/mm/MaXSYZJ22,liu-etal-2022-cross}, improved retrieval methods~\cite{DBLP:conf/icml/BertasiusWT21,DBLP:conf/cvpr/DongLXJH0W19,DBLP:conf/sigir/JinZZZHZ21}, and video-language datasets construction~\cite{DBLP:conf/cvpr/XuMYR16}, it remains challenging to precisely match video and language due to the raw data being in heterogeneous spaces with significant differences.

Current VTR research~\cite{DBLP:journals/ijon/LuoJZCLDL22,DBLP:conf/mm/MaXSYZJ22,DBLP:conf/eccv/LiuXXCJ22} mainly aims to learn a joint feature space across modalities and then compares representations in this space. 
However, with the huge discrepancy between different modalities and the design of modality-independent encoders,
it is challenging to directly compare and calculate the similarities between representations of different modalities generated from different encoders~\cite{liang2022mind}. 
To alleviate the mismatch caused by heterogeneous encoders and data formats, \citet{liu-etal-2022-cross,DBLP:conf/aaai/CaoW0022} proposed to align different modalities in a common space without supervision from text or video. 
However, because of the unsupervised design, the shared spaces are either randomly initialized or updated in an unsupervised fashion, which blocks the power of that aligned space. 
We argue that learning a shared aligned space with supervision is a promising way to improve video-text retrieval. 
Borrowing from text retrieval~\cite{karpukhin-etal-2020-dense,zhao_sparta_2021,gao_coil_2021}, we represent the aligned space and the space containing representations generated by modality-dependent encoders as sparse and dense spaces, respectively, as the aligned space typically carries specific semantics.

In this work, we propose a \textit{\textbf{S}upervised \textbf{S}hared \textbf{S}parse \textbf{M}ulti-grained \textbf{A}lignment framework} for VTR, namely \ours, in which the aligned sparse space is updated under the supervision of the video-text data at hand. 
Specifically, we initialize a finite number of sparse concepts by clustering a large number of basic concepts (words) to form the fine-grained aligned sparse space. 
In return, each sparse concept is composed of several words, which improves the interpretability of our model. 
Then, we match the sparse text and video representations effectively 
by projecting the video representation generated by the video encoder to this fine-grained sparse space. 
The sparse sentence (text) representations can be obtained by looking up the sparse concepts. 
To obtain sparse video representations, we first calculate the cosine similarity between the video representations and the sparse concepts. 
Next, by summing up all the sparse concepts with the weight of the cosine similarity between video representation and sparse concepts, we obtain the sparse video representations.
Furthermore, to better match these two sparse representations, we design two loss functions to update sparse concepts, pushing the sparse representations of text and video as close as possible in the shared sparse space. 
This shared sparse space design not only improves the performance on VTR, but also allows us to interpret what the models have learned. 
The sparse aligned space, as shown in \Cref{fig: examples sparse}, enables the model to accurately capture the key concepts, resulting in improved alignment within the sparse space.

Recently, \citet{DBLP:conf/mm/MaXSYZJ22} demonstrated that incorporating fine-grained video representations~(such as frame or segment representations) with high-level video features can further improve retrieval performance. 
Inspired by their work, we further project \textit{frame} representations into our designed aligned sparse space. 
Compared to high-level video representations, frame representations can be mapped to more detailed concepts, which enriches the overall video representations. 
In this way, we have fine-grained (frame) and coarse-grained (video and sentence) representations from the sparse space and the dense space, enabling us to calculate multi-space multi-grained similarity for exploring the potential of supervised sparse space. 

Finally, to evaluate the effectiveness of our proposed \ours, we conducted experiments on three video-text benchmarks~\cite{chen-dolan-2011-collecting,caba2015activitynet,DBLP:conf/cvpr/XuMYR16}. 
Benefiting from multi-grained and multi-space similarity, our proposed \ours \ outperforms previous methods on all the benchmarks without requiring any additional data during training.

In summary, our contributions are as follows\footnote{The code is released at \href{https://github.com/yimuwangcs/Better_Cross_Modal_Retrieval}{link}.}:
\begin{itemize}
    \item We propose the shared sparse space to alleviate the problem of mismatched representations from different modalities, which arises from the raw data being in heterogeneous spaces and the heterogeneous design of modality-dependent encoders.
    \item Our proposed \ours\ achieves SOTA performance on several metrics across three VTR benchmarks.
    \item Detailed analysis reveals the importance of shared sparse space and multi-grained similarity. Besides, we demonstrate that the design of shared sparse space and multi-grained similarity significantly impacts retrieval performance.
\end{itemize}

\section{Related Works}

\begin{figure*}[t!]
\centering
\includegraphics[width=0.8\textwidth]{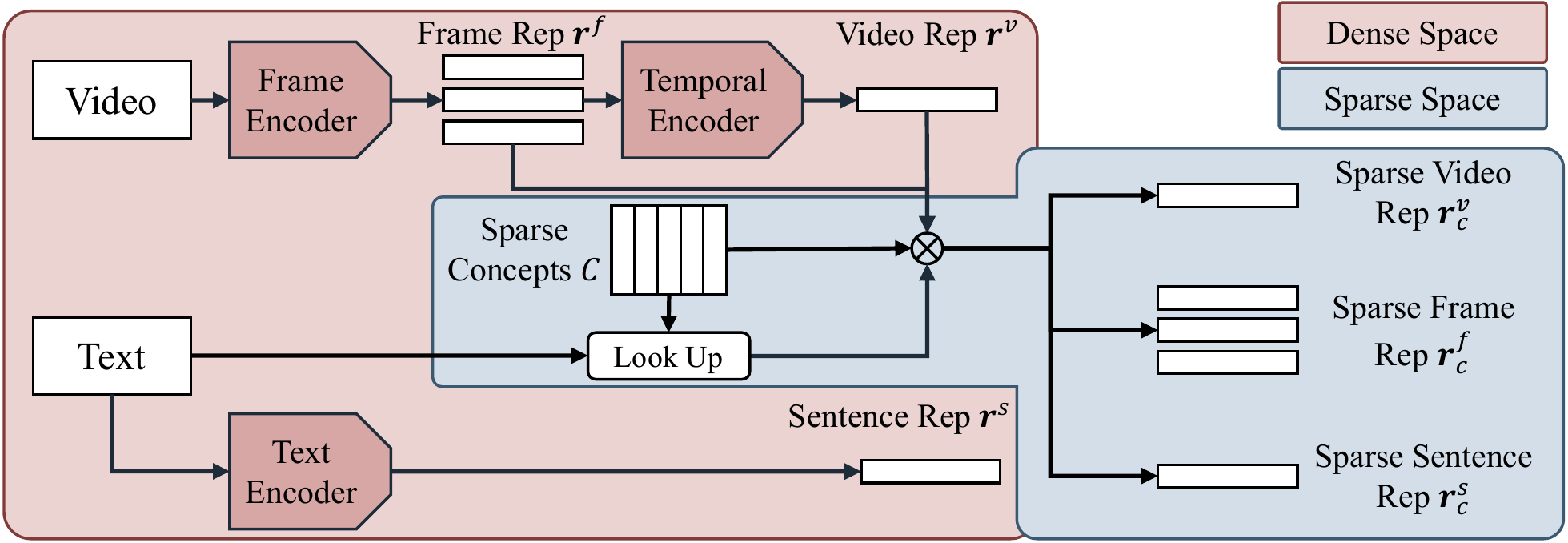} 
\caption{
The illustration of representation generation in our proposed \textit{\textbf{S}upervised \textbf{S}hared \textbf{S}parse \textbf{M}ulti-grained \textbf{A}lignment framework}, namely \ours.
Specifically, for multi-space alignment, we employ a shared sparse space which is consisted of a number of sparse concepts. 
The shared sparse space is updated in a supervised manner during the training procedure, leading to the construction of a fine-grained sparse space. ``$\otimes$'' refers to the calculation in Eqs.~\eqref{eq:sentencecodebook}, \eqref{eq: videocodebook}, and \eqref{eq: framecodebook}.
}\label{fig: representation}
\end{figure*}

Video-Text Retrieval (VTR), which involves cross-modal alignment and abstract understanding of temporal images (videos), has been a popular and fundamental task of language-grounding problems~\cite{10.1145/3394171.3413882,DBLP:conf/prcv/WangWXZ20,ijcai2021p156,yu2023multimodal}. 
Most existing conventional video-text retrieval frameworks~\cite{DBLP:conf/cvpr/YuKCK17,DBLP:conf/cvpr/DongLXJH0W19,DBLP:conf/cvpr/ZhuY20a,DBLP:conf/cvpr/MiechASLSZ20,DBLP:conf/eccv/Gabeur0AS20,DBLP:conf/cvpr/DzabraevKKP21,DBLP:conf/iccv/CroitoruBLJZAL21} focus on learning powerful representations for video and text and extracting separated representations. 
Inspired by the success of self-supervised pretraining methods~\cite{devlin-etal-2019-bert,radford2019language,DBLP:conf/nips/BrownMRSKDNSSAA20} and vision-language pretraining~\cite{DBLP:conf/eccv/Li0LZHZWH0WCG20,DBLP:conf/nips/Gan0LZ0020,DBLP:conf/cvpr/SinghHGCGRK22} on large-scale unlabeled cross-modal data, recent works~\cite{DBLP:conf/cvpr/LeiLZGBB021,DBLP:journals/corr/abs-2109-04290,DBLP:journals/corr/abs-2111-05610,DBLP:conf/mm/MaXSYZJ22,park-etal-2022-exposing,DBLP:conf/mm/WangXHLJHD22,9878037,10.1145/3477495.3531950,DBLP:conf/cvpr/GortiVMGVGY22} have attempted to pretrain or fine-tune video-text retrieval models in an end-to-end manner. 
Frozen in time~\cite{DBLP:conf/iccv/BainNVZ21} uses end-to-end training on both image-text and video-text pairs data by uniformly sampling video frames. 
CLIP4Clip~\cite{DBLP:journals/ijon/LuoJZCLDL22} finetunes models and investigates three similarity calculation approaches for video-sentence contrastive learning on CLIP~\cite{DBLP:conf/icml/RadfordKHRGASAM21}. 
Later, to enable unsupervised sparse learning in VTR, DiscretCodebook~\cite{liu-etal-2022-cross} aligns modalities in a shared space filled with concepts, which are randomly initialized and unsupervisedly updated, while VCM~\cite{DBLP:conf/aaai/CaoW0022} constructs a sparse space with unsupervisedly clustered visual concepts. 
At the same time, OA-Trans~\cite{wang_object-aware_2022} and TABLE~\cite{chen_tagging_2023} both employ a small number of semantic tags as the input to the text encoder to improve alignment between modalities. 

However, due to the unsupervised design, concepts in DiscretCodebook and VCM are either randomly initialized or updated unsupervisedly, which limits the potential of aligned sparse space. 
On the other hand, OA-Trans and TABLE only employ a limited number of concepts to serve as the input of the text encoder to encourage alignment. 
Meanwhile, these methods only perform the \textit{coarse-grained} video-text similarity, lacking the fine-grained contrast between different modalities. 
In comparison, our proposed \ours \ learn the aligned sparse space containing a large number of words in a \textit{supervised} manner, under the supervision of text, and calculate frame-sentence similarity for \textit{multi-space multi-grained} alignment.

\section{Methods}

In this section, we introduce our proposed framework for video-text retrieval, which aligns language and video in a shared sparse space. 
Typically, in video-text retrieval, we have a set of examples $\{(\video_{i}, \text_{i})\}_{i \in [N]}$, where $N$ is the number of examples that are of video and language.

\subsection{General Video-Text Retrieval Paradigm}
In this part, we present a general video-text retrieval framework widely used by previous methods~\cite{DBLP:journals/ijon/LuoJZCLDL22,liu-etal-2022-cross}. 
With this paradigm, we can obtain three representations for different modalities from the dense space, \ie, frame representation $\framefeature$, video representation $\videofeature$, and sentence representation $\sentencefeature$ by modality-dependent encoders.

\smallskip \noindent \textbf{Frame and video representations:} 
Given a video $\video$, several video frames are first sampled as the inputs of the frame encoder to obtain the frame features $\framefeature \in \mathbb{R}^{n_{\text{frame}}\times d}$, where $n_{frame}$ is the number of frames and $d$ is the dimension of features. 
As the frame representations $\framefeature$ are extracted through sampling, to explore the temporal correlation among different frames, we employ a temporal encoder to aggregate frame representations.
With the temporal encoder and the frame representations $\framefeature$, we obtain the video representations $\videofeature \in \mathbb{R}^{1\times d}$.

\smallskip \noindent \textbf{Sentence representation:} 
Given a sentence $\text$, we use a text encoder to obtain the text representation $\sentencefeature \in \mathbb{R}^{1\times d}$.

\subsection{Fine-Grained Aligned Sparse Space}

The key to the video-text retrieval task is to precisely align representations from different modalities. 
However, due to the heterogeneous encoder architectures and data formats of different modalities, it is difficult to align directly~\cite{liang2022mind}. 
Therefore, instead of directly enforcing the representations to be aligned, we propose aligning them in an aligned sparse constructed by $n_{c}$ sparse concepts $C \in \mathbb{R}^{n_{c} \times d}$.
Each sparse concept $\mathbf{c}$ represents several basic concepts (words).
Moreover, to supervise the updates of sparse concepts, we utilize the human-annotated knowledge at hand, \ie, text annotations in the paired video-text data.

\smallskip \noindent 
\textbf{Initialization}. 
First, we map all the words into embeddings by the embedding layer $f_{emb}$ of the text encoder. 
But as the number of words is relatively large (for example, in Clip~\cite{DBLP:conf/icml/RadfordKHRGASAM21}, the number of sub-words is approximately 30k), we cluster embeddings into $n_{c}$ clusters using KNN~\cite{4447274} to form the sparse concepts $C$ and represent all the words by their cluster's centers $\mathbf{c}$. 
Consequently, each sparse concept $\mathbf{c}$ represents a bunch of words that are similar on the embedding space, enabling fine-grained alignment. 
The mapping from words to sparse concepts is denoted by $h_{w2c} \in [n_{words}] \rightarrow \{0,1\}^{n_{c} \times 1}$. 
Now, $n_{c}$ sparse concepts have been initialized. 

\smallskip \noindent 
\textbf{Obtaining the sparse sentence representation}. 
For text, as the caption is at hand, we can directly 
tokenize the sentences into words and look up the corresponding sparse concepts in $C$.
The sparse sentence representation $\sentencecodebook \in \mathbb{R}^{1 \times d}$ is obtained by averaging all the representations of concepts that are fetched with the surface form of the sentence, as follows,
\begin{equation}
    \sentencecodebook = sim^{\text\top} C / |\text|\,,\label{eq:sentencecodebook}
\end{equation}
where $|\text|$ is the number of words in $\text$ and $sim^{\text} = \sum_{\word \in \text}h_{w2c}(\word)$ is a vector with the length of $n_{c}$.

\smallskip \noindent 
\textbf{Obtaining the sparse video representation}. 
We first calculate the cosine similarity $\videocodebooksim \in \mathbb{R}^{1\times n_{c}}$ between the video representations and sparse concepts $C$ as $\videocodebooksim_{j} = \operatorname{cos}(\videofeature, \mathbf{c}_{j}), \forall j \in [n_c]$, where $\videocodebooksim_{j}$ is the $j$-th element of $\videocodebooksim$ and $\operatorname{cos}(\cdot, \cdot)$ is the cosine similarity. 
Next, sparse video representations are obtained by weighted summing the sparse concepts as,
\begin{align}
    \label{eq: videocodebook}\videocodebook &= \videocodebooksim C / \|\videocodebooksim\|_1 \,.
\end{align}

\smallskip \noindent 
\textbf{Obtaining the sparse frame representation}. 
Similarly, the cosine similarity $\framecodebooksim \in$$ \mathbb{R}^{n_{frame} \times n_{c}}$ between the frame representations and sparse concepts is calculated as $\framecodebooksim_{i,j} = \operatorname{cos}(\framefeature_{i}, \mathbf{c}_{j}), \forall i \in [n_{frame}], \forall j \in [n_{c}]$, where $\framecodebooksim_{i,j}$ is the $(i,j)$-th element of $\framecodebooksim$ and $\framefeature_{i}$ is the $i$-th row of $\framefeature$. 
Next, sparse frame representations are obtained as,
\begin{align}
    \framecodebook &= \sum_{i \in [n_{frame}]} \framecodebooksim_{i} C / \|\framecodebooksim_{i}\|_1\label{eq: framecodebook}\,.
\end{align}

Finally, we have the sparse frame, video, and sentence representations $\framecodebook \in \mathbb{R}^{n_{frame} \times d}, \videocodebook \in \mathbb{R}^{1 \times d}, \sentencecodebook \in \mathbb{R}^{1 \times d}$ with the frame and video sparse space similarity $\framecodebooksim \in \mathbb{R}^{n_{frame} \times n_{c}}$ and $\videocodebooksim \in \mathbb{R}^{n_{c}}$ along with the sentence sparse space similarity (supervision) $sim^{\text}$.

\begin{figure}[t!]
\centering
\includegraphics[width=1\columnwidth]{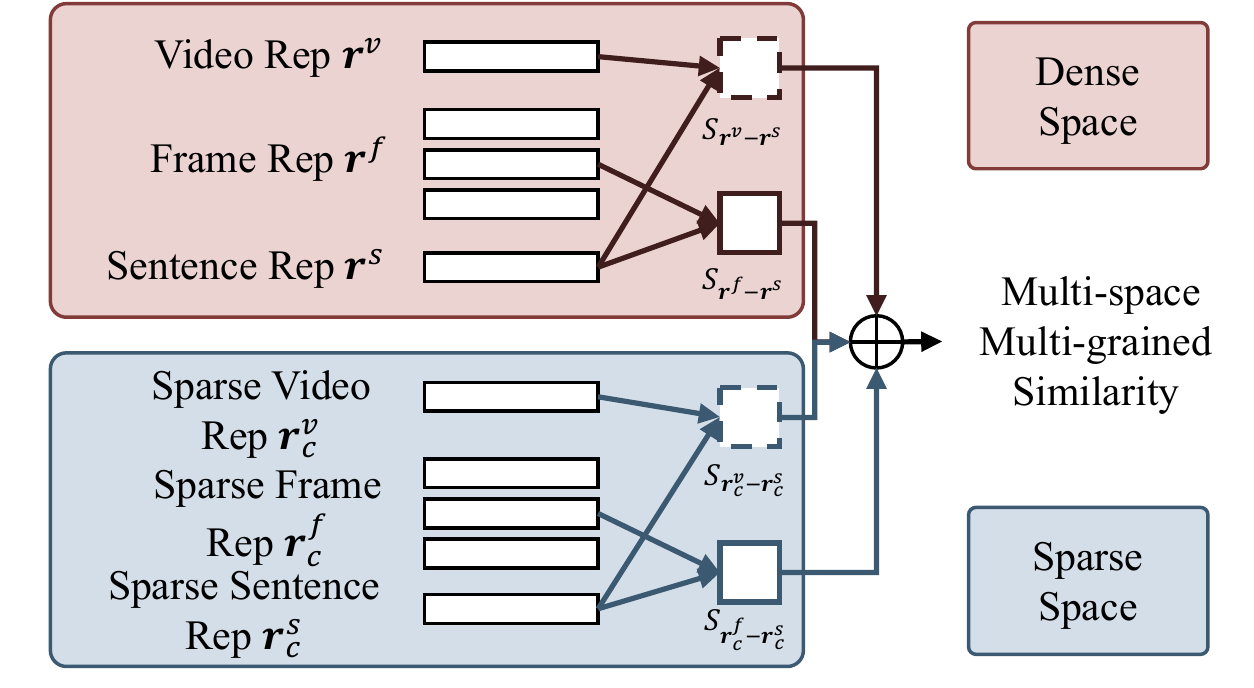}
\caption{
The illustration of similarity calculation. 
To enable multi-space multi-grained alignment, we calculate fine-grained (frame-sentence) and coarse-grained (video-sentence) similarity. Our preliminary experiments showed that the text encoder has a good ability to capture semantics, so we only use sentence representations for the text modality.
}\label{fig: sim_cal}
\end{figure}

\subsection{Multi-Space Multi-Grained Similarity}

In this part, we will demonstrate our method for calculating the similarities between data from two different modalities, as shown in \Cref{fig: sim_cal}, 
including the similarities in the dense space and in shared sparse space, inspired by \citet{DBLP:conf/mm/MaXSYZJ22}. 
We can now compute multi-space (sparse and dense spaces) multi-grained (fine-grained and coarse-grained) similarity for precise alignment.

\subsubsection{Dense Space Similarity}

\textbf{Video-Sentence similarity $\videotosentencesim$}. 
To obtain a fine-grained similarity, we use a learnable matrix $\videotosentenceattention \in \mathbb{R}^{d\times d}$ to focus on the discriminative features of video and sentence representations as,
\begin{equation*}
    \videotosentencesim =  \videofeature \videotosentenceattention \feature^{s\top}\,.
\end{equation*}
\smallskip \noindent 
\textbf{Frame-Sentence similarity $\frametosentencesim$}. 
To obtain a fine-grained similarity, we first calculate an \textit{instance-aware weight} using the softmax function applied to the dot product of $\sentencefeature \feature^{f\top}$, and then use a learnable matrix $\frametosentenceattention \in \mathbb{R}^{n_{frame} \times n_{frame}}$ to focus on discriminative frames. 
In this way, the similarity is calculated as,
\begin{equation*}
    \frametosentencesim = \operatorname{softmax}(\sentencefeature \feature^{f\top}) \frametosentenceattention \framefeature \feature^{s\top}\,.
\end{equation*}

\subsubsection{Sparse Space Similarity}

\textbf{Video-Sentence shared sparse space similarity $\videotosentencecodebooksim$}. 
Similarly, to obtain a fine-grained similarity on the shared sparse space, we use a learnable matrix $\videotosentencecodebookattention \in \mathbb{R}^{d\times d}$ to focus on the discriminative features of sparse video and sentence representations. 
Now, the similarity is calculated as,
\begin{equation*}
    \videotosentencecodebooksim = \videocodebook \videotosentencecodebookattention \feature^{s\top}_{c}\,.
\end{equation*}

\smallskip \noindent 
\textbf{Frame-Sentence shared sparse space similarity $\frametosentencecodebooksim$}. 
With \textit{instance-aware weights} $\operatorname{softmax}(\sentencecodebook \feature^{f\top}_{c})$ and a learnable matrix $\frametosentencecodebookattention \in \mathbb{R}^{n_{frame} \times n_{frame}}$, we get the similarity between the sparse frame and sentence representations as,
\begin{equation*}
    \frametosentencecodebooksim = \operatorname{softmax}(\sentencecodebook \feature^{f\top}_{c}) \frametosentencecodebookattention \framecodebook \feature^{s\top}_{c}\,.
\end{equation*}

\subsubsection{Overall Similarity}
The overall video-text similarity is defined as,
\begin{equation*}
    \sim = \frac{\frametosentencesim + \videotosentencesim + \frametosentencecodebooksim + \videotosentencecodebooksim}{4}\,.
\end{equation*}

\subsection{Objective}

The objective consists of three different losses. 
The first component is contrastive loss.
Following Clip4Clip~\cite{DBLP:journals/ijon/LuoJZCLDL22}, we employ the symmetric InfoNCE loss over the similarity matrix to optimize the retrieval model as,
\begin{align*}
    \ell_{sim} = & \ell_{v2t} + \ell_{t2v} \\
    = & - \frac{1}{N} \sum_{i \in [N]} \log \frac{\exp(\sim_{i,i})}{\sum_{j\in[N]} \exp(\sim_{i,j})}\\
    & - \frac{1}{N} \sum_{i \in [N]} \log \frac{\exp(\sim_{i,i})}{\sum_{j\in[N]} \exp(\sim_{j,i})}\,,
\end{align*}
where $\sim_{i,j}$ is similarity between $i$-th video and $j$-th text and $N$ is the number of paired data.

The second loss we minimize is the alignment loss, which matches the sparse frame and video representations ($\framecodebook$ and $\videocodebook$) with the sparse sentence representations $\sentencecodebook$ in the $\ell_2$ distance, as,
\begin{align*}
    \ell_{align} = & \frac{1}{N} \sum_{i \in [N]} \left( \left\|\videocodebook - \sentencecodebook \right\|_2 \right. \\
    &\left. + \left\| \frac{\mathbf{1} \mathbf{r}_{c}^{f}}{n_{frame}}  - \sentencecodebook\right\|_2\right)\,,
\end{align*}
where $\mathbf{1}$ is the vector only containing $1$.

In addition, to match the frame and video representations with the corresponding sparse concepts, we minimize the sparse similarity loss as, 
\begin{align*}
    \ell_{sparse} = & \frac{1}{N} \sum_{i \in [N]} \left( \left\|\videocodebooksim - \sentencecodebooksim \right\|_2 \right. \\
    & \left.+  \left\| \frac{\mathbf{1}\framecodebooksim}{n_{frame}}  - \sentencecodebooksim \right\|_2\right)\,,
\end{align*}

The overall objective is the linear combination of the above three losses as,
\begin{equation*}
    \ell = \ell_{sim} + \alpha \ell_{align} + \beta \ell_{sparse}\,,
\end{equation*}
where $\alpha$ and $\beta$ are hyperparameters controlling the trade-off between three losses. 
We set $\alpha=0.02$ and $\beta =0.01$ for all the experiments.

\begin{table*}[ht!]
\centering
\resizebox{\textwidth}{!}{%
\begin{tabular}{ll|ccccc|ccccc}
\toprule
 \multirow{2}{*}{Methods}  & \multirow{2}{*}{Venue}          & \multicolumn{5}{c|}{Text-to-Video Retrieval}                                        & \multicolumn{5}{c}{Video-to-Text Retrieval}                                        \\
        &     & R@1$\uparrow$ & R@5$\uparrow$ & R@10$\uparrow$ & MdR$\downarrow$ & MnR$\downarrow$ & R@1$\uparrow$ & R@5$\uparrow$ & R@10$\uparrow$ & MdR$\downarrow$ & MnR$\downarrow$ \\
\midrule
VLM & ACL'21 & 28.1 & 55.5 & 67.4 & 4.0 & - & -&-&-&-&-\\
HERO & EMNLP'21 & 16.8 & 43.3 & 57.7 & - & -&-&-&-&- & -\\
VideoCLIP &EMNLP'21 &  30.9 & 55.4 & 66.8 & - & -&-&-&-&- & -\\
EvO & CVPR'22 & 23.7 & 52.1 & 63.7& 4.0 & - & -&-&-&-&-\\
OA-Trans & CVPR'22 & 35.8 & 63.4 & 76.5 & 3.0 & - & -&-&-&-&-\\
RaP & EMNLP'22 & 40.9 & 67.2 & 76.9 & 2.0 & -&-&-&-&-\\
\rowcolor{gray!10}\multicolumn{12}{l}{\textit{BLIP-based}} \\ 
LiteVL-S & EMNLP'22 & 46.7 & 71.8 & 81.7 & 2.0 & - & -&-&-&-&-\\
\midrule
\rowcolor{gray!10}\multicolumn{12}{l}{\textit{ViT-B/32-based}}\\
Align\&Tell  & TMM       & 45.2          & 73.0          & 82.9           & 2.0             & -               & 43.4          & 70.9          & 81.8           & 2.0             & -               \\
X-Pool    & CVPR'22 & 46.9          & 72.8          & 82.2           & 2.0               & 14.3            & -             & -             & -              & -               & -               \\
CenterCLIP  & SIGIR'22  & 44.2          & 71.6          & 82.1           & 2.0               & 15.1            & {42.8}          & {71.7}          & {82.2}           & 2.0               & {10.9}            \\
TS2-Net  & ECCV'22  & {47.0}          & \underline{74.5}          & \underline{83.8}           & 2.0             & \underline{13.0}            & 45.3          & {74.1}          & {83.7}           & 2.0             & {9.2}             \\
X-CLIP   & ACM MM'22  & 46.1          & {74.3}          & {83.1}           & 2.0             & {13.2}            & {46.8}          & 73.3          & {84.0}           & 2.0             & \underline{9.1}             \\
NCL    & EMNLP'22 & 43.9 & 71.2 & 81.5 & 2.0 & 15.5 & 44.9 & 71.8 & 80.7 & 2.0 & 12.8      \\
TABLE & AAAI'23 & 47.1 & 74.3 & 82.9 & 2.0 & 13.4 & 47.2 & \underline{74.2} & \underline{84.2} & 2.0 & 11.0\\
VOP & CVPR'23 & 44.6 & 69.9 & 80.3 & 2.0 & 16.3 & 44.5 & 70.7 & 80.6 & 2.0 & 11.5 \\
\midrule
CLIP4Clip & NC & 44.5          & 71.4          & 81.6           & 2.0               & 15.3            & -             & -             & -              & -               & -               \\
DiscreteCodebook& ACL'22   & 43.4          & {72.3}          & 81.2           & -               & {14.8}            & 42.5          & 71.2          & 81.1           & -               & 12.0            \\
VCM         & AAAI'22        & 43.8          & 71.0          & -           & 2.0             & 14.3            & 45.1          & 72.3          & 82.3           & 2.0             & 10.7            \\
\rowcolor{green!10}\ours  &    & \underline{49.1}   & 73.9   & 82.8   & 2.0    & 13.5    & \underline{46.9}   & {73.8}   & 82.1   & 2.0    & 9.3             \\
\rowcolor{green!10} $\ours^{\dag}$  &    & \textbf{51.7}   & \textbf{75.9}   & \textbf{85.4}   & 1.0    & \textbf{11.1}    & \textbf{51.6}   & \textbf{76.8}   & \textbf{85.0}   & 1.0    & \textbf{8.4}             \\
\bottomrule
\rowcolor{gray!10}\multicolumn{12}{l}{\textit{ViT-B/16-based}}\\
Align\&Tell  & TMM        & 47.4          & 74.3          & {84.1}           & 2.0             & -               & 45.3          & 73.5          & 83.7           & 2.0             & -               \\
CenterCLIP  & SIGIR'22  & 48.4          & 73.8          & 82.0           & 2.0               & 13.8            & \underline{47.7}          & 75.0          & 83.3           & 2.0               & {10.2}            \\
HiSE    & ACM MM'22   & 45.0          & 72.7          & 81.3           & 2.0             & -               & 46.6          & 73.3          & 82.3           & 2.0             & -               \\
TS2-Net  & ECCV'22  & {49.4}          & \underline{75.6}          & \underline{85.3}           & 2.0             & {13.5}            & 46.6          & {75.9}          & \underline{84.9}           & 2.0             & \underline{8.9}             \\
\midrule
CLIP4Clip & NC & 45.8*         & 74.3*         & {84.1}*          & 2.0*              & -               & 43.2*         & 71.3*         & 82.0*          & 2.0*              & -               \\
\rowcolor{green!10}\ours  &    & \underline{49.8}          & {75.1}          & 83.9           & 2.0             & \underline{12.2}            & {47.3}          & \underline{76.0}          & {84.3}           & 2.0             & \underline{8.9}            \\
\rowcolor{green!10}$\ours^{\dag}$  &    & \textbf{53.1}          & \textbf{78.2}          & \textbf{86.2}           & 1.0             & \textbf{10.5}            & \textbf{52.7}          & \textbf{79.2}          & \textbf{86.3}           & 1.0             & \textbf{8.2}            \\
\bottomrule
\end{tabular}%
}
\caption{Video-Text retrieval results on MSR-VTT. * represents data copied from Align\&Tell. The best results are marked in \textbf{bold}. The second best results are \underline{underlined}. ``NC'' refers to Neurocomputing. $\dag$ refers to the results with the inverted softmax.
}
\label{tab:msr-vtt}
\end{table*}

\section{Experiments}

\subsection{Datasets and Baselines}

To show the empirical efficiency of our \ours, we train it on MSR-VTT~\cite{DBLP:conf/cvpr/XuMYR16}, MSVD~\cite{chen-dolan-2011-collecting}, and ActivityNet~\cite{caba2015activitynet}. 
We compare with 
VLM~\cite{xu-etal-2021-vlm}, 
HERO~\cite{li-etal-2020-hero},
VideoCLIP~\cite{xu-etal-2021-videoclip},
EvO~\cite{shvetsova_everything_2022},
OA-Trans~\cite{wang_object-aware_2022},
RaP~\cite{wu-etal-2022-rap},
LiteVL~\cite{chen-etal-2022-litevl},
NCL~\cite{park-etal-2022-normalized},
TABLE~\cite{chen_tagging_2023},
VOP~\cite{huang_vop_2023},
Clip4Clip~\cite{DBLP:journals/ijon/LuoJZCLDL22}, 
X-CLIP~\cite{DBLP:conf/mm/MaXSYZJ22}, 
DiscreteCodebook~\cite{liu-etal-2022-cross}, 
TS2-Net~\cite{DBLP:conf/eccv/LiuXXCJ22}, 
VCM~\cite{DBLP:conf/aaai/CaoW0022}, 
HiSE~\cite{DBLP:conf/mm/WangXHLJHD22}, 
Align\&Tell~\cite{9878037}, 
CenterCLIP~\cite{10.1145/3477495.3531950}, 
and
X-Pool~\cite{DBLP:conf/cvpr/GortiVMGVGY22}. 
Implementation details and evaluation protocols are deferred to the Appendix.

\begin{table}[t!]
\centering
\resizebox{\columnwidth}{!}{%
\begin{tabular}{ll|cccc}
\toprule
                    \multirow{2}{*}{Methods}  & \multirow{2}{*}{Venue}& \multicolumn{4}{c}{Text-to-Video Retrieval}                      \\
                    &           & R@1$\uparrow$ & R@5$\uparrow$ & R@10$\uparrow$ & MnR$\downarrow$ \\
                               \midrule
\multicolumn{6}{c}{\textit{MSVD}}\\\midrule
X-CLIP    & ACM MM'22  & 47.1          & \underline{77.8}          & -              & \underline{9.5}             \\
HiSE            & ACM MM'22               & 45.9          & 76.2          & 84.6           & -               \\
X-Pool       & CVPR'22                  & \underline{47.2}          & 77.4          & \textbf{86.0}           & \textbf{9.3}             \\
\midrule
CLIP4Clip & NC & 45.2          & 75.5          & 84.3           & 10.3            \\
\rowcolor{green!10}\ours  &                        & \textbf{47.3}          & \textbf{78.8}          & \underline{85.7}           & \textbf{9.3}             
\\\midrule
\multicolumn{6}{c}{\textit{ActivityNet}}\\\midrule
Align\&Tell   & TMM     & 42.6          & 73.8          & -              & -               \\
X-CLIP    & ACM MM'22          & \underline{44.3}          & \underline{74.1}          & -              & 7.9             \\
TS2-Net  & ECCV'22                      & 41.0          & 73.6          & \underline{84.5}           & 8.4             \\
\midrule
CLIP4Clip &NC & 40.5          & 72.4          & -               & 7.5             \\
VCM        & AAAI'22                    & 40.8          & 72.8          & -              & \underline{7.3}             \\
\rowcolor{green!10}\ours     &       & \textbf{45.0}          & \textbf{75.5}          & \textbf{85.7}           & \textbf{6.3}            \\
\bottomrule
\end{tabular}%
}
\caption{Text-Video retrieval results on MSVD and ActivityNet. The best results are marked in \textbf{bold}. The second best results are \underline{underlined}.}
\label{tab:msvd}
\end{table}

\begin{table*}[h!]
\centering
\resizebox{\textwidth}{!}{%
\begin{tabular}{l|ccccc|ccccc}
\toprule
             & \multicolumn{5}{c|}{Text-to-Video Retrieval}                                        & \multicolumn{5}{c}{Video-to-Text Retrieval}                                        \\
             & R@1$\uparrow$ & R@5$\uparrow$ & R@10$\uparrow$ & MdR$\downarrow$ & MnR$\downarrow$ & R@1$\uparrow$ & R@5$\uparrow$ & R@10$\uparrow$ & MdR$\downarrow$ & MnR$\downarrow$ \\
\midrule
\ours\ (ViT-B/32) w. SE                  &   47.3 & 73.5 & 82.0 & 2.0 & 13.3 & 45.6 & 73.4 & \textbf{82.4} & 2.0 & \textbf{9.1}   \\
\rowcolor{green!10}\ours\ (ViT-B/32) w. Emb     & \textbf{49.1}   & \textbf{73.9}   & \textbf{82.8}  & 2.0    & \textbf{13.5}    & \textbf{46.9}   & \textbf{73.8}   & {82.1}   & 2.0    & 9.3             \\

\bottomrule
\end{tabular}%
}
\caption{
Comparing the power of different sparse spaces on MSR-VTT. 
``Emb'' and ``SE''  refers to the embedding space and semantic embedding space.
}
\label{tab: semantic embeddings}
\end{table*}

\begin{table*}[h!]
\centering
\resizebox{\textwidth}{!}{%
\begin{tabular}{l|ccccc|ccccc}
\toprule
             & \multicolumn{5}{c|}{Text-to-Video Retrieval}                                        & \multicolumn{5}{c}{Video-to-Text Retrieval}                                        \\
             & R@1$\uparrow$ & R@5$\uparrow$ & R@10$\uparrow$ & MdR$\downarrow$ & MnR$\downarrow$ & R@1$\uparrow$ & R@5$\uparrow$ & R@10$\uparrow$ & MdR$\downarrow$ & MnR$\downarrow$ \\
\midrule
\ours\ (ViT-B/32) w/o clustering                  &  48.7 & \textbf{74.4}& \textbf{83.0} & 2.0 & \textbf{13.4} & 46.7 & 73.3& \textbf{82.6} & 2.0 & \textbf{9.2} \\
\rowcolor{green!10}\ours\ (ViT-B/32)      & \textbf{49.1}   & 73.9   & 82.8  & 2.0    & 13.5    & \textbf{46.9}   & \textbf{73.8}   & 82.1   & 2.0    & 9.3             \\
\bottomrule
\end{tabular}%
}
\caption{Ablation study on the effect of clustering when constructing the shared sparse space.}
\label{tab: not cluster}
\end{table*}
\begin{table}[t!]
\centering
\resizebox{\columnwidth}{!}{%
\begin{tabular}{l|ccc|ccc}
\toprule
\multirow{2}{*}{Size} & \multicolumn{3}{c|}{Text-to-Video Retrieval}                          & \multicolumn{3}{c}{Video-to-Text Retrieval}   \\
                               & R@1                     & R@5              & MnR & R@1                     & R@5              & MnR \\
                               \midrule
512                            & \underline{48.7}        & 73.0             & \textbf{12.9}  & 46.4                    & 72.8             & \textbf{9.0}   \\
\rowcolor{green!10}1024                           & \textbf{49.1}           & \textbf{73.9}    & \underline{13.5}  & \underline{46.9}                    & \textbf{73.8}    & 9.3   \\
2048                           & 48.3                    & \textbf{73.9}    & \underline{13.5}  & \textbf{47.0}                    & 72.7             & \underline{9.1}   \\
4096                           & 47.6                    & \underline{73.6} & 13.6  & 46.8                    & \underline{73.4} & 9.3   \\ \midrule
DC (1024)    & 43.4          & 72.3                   & 14.8            & 42.5          & 71.2               & 12.0            \\
VCM                  & 43.8          & 71.0                & 14.3            & 45.1          & 72.3             & 10.7            \\
\bottomrule
\end{tabular}%
}
\caption{Retrieval performance with different sizes of sparse space on the MSR-VTT dataset using \ours\ with ViT/B-32. ``DC'' represents DiscreteCodebook~\cite{liu-etal-2022-cross}, which also aligns modalities in a sparse space whose size is 1024 with the base model of ViT/B-32. The best results are marked in \textbf{bold}. The second best results are \underline{underlined}.}
\label{tab:ablation codebook size}
\end{table}

\begin{table*}[ht!]
\centering
\resizebox{\textwidth}{!}{%
\begin{tabular}{cc|cc|ccccc|cccccc}
\toprule
 \multicolumn{2}{c|}{Dense Space} & \multicolumn{2}{c|}{Sparse Space} & \multicolumn{5}{c|}{Text-to-Video Retrieval}                                          & \multicolumn{5}{c}{Video-to-Text Retrieval}                                                                                     \\
                           S-V    & S-F   & S-V       & S-F      & R@1$\uparrow$ & R@5$\uparrow$ & R@10$\uparrow$ & MdR$\downarrow$ & MnR$\downarrow$ & R@1$\uparrow$ & R@5$\uparrow$ & R@10$\uparrow$ & MdR$\downarrow$ & MnR $\downarrow$ \\\midrule
 \Checkmark      &                  &                      &                                          & 42.8          & 72.0          & 82.3           & 2.0             & 15.0              & 41.9          & 71.1          & 81.5           & 2.0             & 11.1                                                             \\
  \Checkmark      &                  & \Checkmark     &    & 43.3          & 70.5          & 81.4           & 2.0             & 15.6              & 42.5          & 71.0          & 80.9           & 2.0             & 11.9                                         \\

                                             & \Checkmark     &                      &                     &  44.4          & 71.8          & 81.8           & 2.0             & 14.5              & 44.1          & 71.8          & 81.7           & 2.0             & 10.4                                                       \\
                                                                                                                    & \Checkmark     &                      & \Checkmark        & 44.8          & 72.1          & 81.7           & 2.0             & 15.9              & 41.7          & 70.2          & 79.6           & 2.0             & 10.8                                                          \\

                           \midrule
                         \Checkmark      &                  &                      & \Checkmark        & 42.9          & 72.3          & 81.6           & 2.0             & 15.2              & 42.0          & 70.9          & 81.1           & 2.0             & 11.0                                                         \\
                                             & \Checkmark     & \Checkmark         &                     & 43.8          & 72.1          & 82.3           & 2.0             & 14.7              & 41.5          & 70.6          & 80.3           & 2.0             & 9.8                                                          \\
                                             \midrule
                                                                         \Checkmark      & \Checkmark     &                      &                     & 44.0          & 71.3          & 80.9           & 2.0             & 14.8              & 43.6          & 69.5          & 80.1           & 2.0             & 10.4                                                         \\

                           \Checkmark      & \Checkmark     & \Checkmark         &                     & 47.4          & 73.3          & 82.4           & 2.0             & \textbf{12.9}              & 46.4          & 73.0          & \textbf{82.2}           & 2.0             & \textbf{8.9}                                                          \\
                           \Checkmark      & \Checkmark     &                      & \Checkmark        & 47.4          & 73.6          & 82.5           & 2.0             & 13.2              & \textbf{47.3}          & 72.3          & 81.7           & 2.0             & \textbf{8.9}                                                          \\
\rowcolor{green!10}                           \Checkmark      & \Checkmark     & \Checkmark         & \Checkmark        &      \textbf{49.1}   & \textbf{73.9}   & \textbf{82.8}   & 2.0    & 13.5    & 46.9   & \textbf{73.8}   & 82.1   & 2.0    & 9.3         \\
                          \bottomrule
\end{tabular}%
    }
\caption{Retrieval performance with different similarities on MSR-VTT using \ours\ with the base model of ViT-B/32. ``S-V'' and ``S-F'' represent Sentence-Video (coarse-grained) and Sentence-Frame (fine-grained) similarities.}
\label{tab:ablation study similarity}
\end{table*}

\begin{table}[t!]
\centering
\resizebox{\columnwidth}{!}{%
\begin{tabular}{lc|ccc|ccc}
\toprule
\multirow{2}{*}{Base Model}           & \multirow{2}{*}{TE} & \multicolumn{3}{c|}{Text-to-Video} & \multicolumn{3}{c}{Video-to-Text} \\
                                 &                     & R@1       & R@5       & MnR      & R@1       & R@5       & MnR     \\ 
                                 \midrule
\multirow{2}{*}{ViT-B/32} &                     & 47.0      & \textbf{73.9}      & 14.5       & 45.7      & 72.3      & 9.6       \\
                        & \cellcolor{green!10}\Checkmark          & \cellcolor{green!10}\textbf{49.1}      & \cellcolor{green!10}\textbf{73.9}      & \cellcolor{green!10}\textbf{13.5}       & \cellcolor{green!10}\textbf{46.9}      & \cellcolor{green!10}\textbf{73.8}      & \cellcolor{green!10}\textbf{9.3}       \\
                                 \midrule
\multirow{2}{*}{ViT-B/16} &   &   47.3&74.9&12.8&46.1&75.1&9.5      \\
 &  \cellcolor{green!10}\Checkmark         & \cellcolor{green!10}\textbf{49.8}      & \cellcolor{green!10}\textbf{75.1}      & \cellcolor{green!10}\textbf{12.2}       & \cellcolor{green!10}\textbf{47.3}      & \cellcolor{green!10}\textbf{76.0}      & \cellcolor{green!10}\textbf{8.9}       \\ 
\bottomrule
\end{tabular}%
}
\caption{Retrieval performance with or without the temporal encoder (``TE'') and with different base models.}
\label{tab:temporal encoder and bigger model}
\end{table}

\subsection{Quantitative Results}

\smallskip \noindent 
\textbf{MSR-VTT.} 
As shown in Table~\ref{tab:msr-vtt}, \ours\ achieves the best R@1 on the text-to-video retrieval results using ViT-B/32 and ViT-B/16, outperforming the second-best method by $2.1$ and $0.4$, respectively. 
The performance of \ours~on the video-to-text retrieval task is also comparable with previous methods, achieving the best and second-best results on R@1 and R@5 using ViT-B/32. 
Moreover, we notice that only 1 previous method using ViT-B/16 outperforms \ours \ with ViT-B/32 on the text-to-video retrieval, demonstrating the effectiveness of \ours. 
Compared to DiscreteCodebook~\cite{liu-etal-2022-cross}, which aligns modalities in an unsupervised manner, \ours\ outperforms DiscreteCodebook on every metric. 
Meanwhile, \ours\ also outperforms VCM~\cite{DBLP:conf/aaai/CaoW0022}, which constructs an aligned space with unsupervisedly clustered visual concepts, demonstrating the importance of supervising alignment in the sparse space. 
This suggests that aligning modalities with fine-grained supervision is a promising approach to improving video-to-text retrieval performance.

\smallskip \noindent 
\textbf{MSVD and ActivityNet.}
The results on MSVD and ActicityNet are shown in \Cref{tab:msvd}.
\ours\ achieves the best R@1 on text-to-video retrieval on two datasets compared to the previous methods. 
Besides, with the shared sparse space and multi-grained alignment, \ours\ also has the lowest MnR.

\subsection{Ablation Studies}
In this part, we present a series of ablation experiments on MSR-VTT to demonstrate the effectiveness of different components of \ours.
The evaluation of two proposed losses, similarity calculation, and the importance of word-level features are deferred to the Appendix.

\subsubsection{Efficiency of Sparse Space}

\textbf{The choice of different initialization of sparse spaces}. 
To choose the best initialization method for the sparse space, we conduct experiments using two different initializations, \ie, the embedding and semantic embedding spaces, as shown in \Cref{tab: semantic embeddings}.
The embedding space is the one we use in \ours, while the semantic embedding space, is initialized by outputs of the last layer in the text encoder, with input consisting of a word and two [SEP] tokens. 
By replacing the embedding initialization with the semantic embedding, the retrieval performance of \ours\ decreases, proving the superiority of embedding space over the semantic embedding space. 

\smallskip \noindent 
\textbf{Size of sparse space. }
Another important factor to consider is the size of the sparse space.
When we have unlimited data to train models, a large sparse space is ideal. 
However, when the data is limited, a large sparse space can lead to sparse gradients, resulting in most of the concepts not being able to be updated, while a small sparse space will restrict the retrieval ability as it becomes more challenging to distinguish between numerous data points. 
The results of these experiments can be found in Table~\ref{tab:ablation codebook size}. 
We see that halving and doubling the size of the sparse space slightly decreases performance.

\smallskip \noindent 
\textbf{Impact of clustering}. 
As \ours~clusters all the embeddings to initialize concept clusters, it is uncertain whether clustering will hinder the power of the shared sparse space. 
Clustering can be useful to extract high-level abstract concepts and reduce noise. 
However, it may also lead to a loss of information, which is important for fine-grained alignment. 
Specifically, we compare the performance of \ours\ to that of a modified version, \ours\ w/o clustering concepts, which directly uses over 30k basic concepts to form the shared sparse space. 
Quantitative results can be found in Table~\ref{tab: not cluster}.
The results show that without clustering, R@5, R@10, and MnR on text-to-video retrieval and R@10 and MnR on video-to-text retrieval are improved.
On one hand, similar basic concepts can be better separated, which leads to more precise alignment. 
On the other hand, that may lead to sparse gradients, resulting in some concepts not being fully updated while others are over-updated. 
This might cause some concepts to be under or over-represented, which might negatively impact the performance~\cite{DBLP:journals/jmlr/RadovanovicNI10}. 
Therefore, it's important to find the balance in clustering to achieve the best performance.

\subsubsection{Efficiency of Multi-Grained Similarities}
In order to fully evaluate the impact of multi-grained similarities, we compare different variants of \ours\ and the results are shown in \Cref{tab:ablation study similarity}. 
From these results, we can draw three conclusions,
\begin{itemize}
    \item Multi-grained similarities are crucial for retrieval. 
    Using both coarse- and fine-grained alignments in the dense space improved R@1 from 42.8 and 41.9 to 44.0 and 43.6 on text-to-video and video-to-text retrieval compared with only using coarse-grained alignment in the dense space, respectively. 
    The same observation can be observed in the sparse space.
    \item 
    Sparse space plays a crucial role in improving the alignment of modalities. 
    We observe that incorporating coarse-grained in the dense and sparse spaces improves R@1 for text-to-video retrieval from 42.8 to 43.3 compared to only performing coarse-grained similarity in the dense space, respectively. 
    \item 
    Using multi-space and multi-grained similarities simultaneously achieves the best performance. 
    R@1 on text-to-video and video-to-text retrieval is significantly improved from 42.8 and 41.9 to 49.1 and 46.9, respectively. 
\end{itemize}

\subsubsection{Temporal Encoder and Larger Model} 
We also investigate the effect of the temporal encoder (TE, a small sequence transformer) and different base models. 
The results are shown in Table~\ref{tab:temporal encoder and bigger model}. 
\ours\ with TE outperforms \ours \ without TE, because it is able to better model the temporal relation among different frames in a video. 
Besides, using a larger base model, such as ViT-B/16, further improves the performance of \ours, as a larger base model typically has better representation learning abilities benefiting this retrieval task as well.
Similar conclusions can be found in previous works \cite{DBLP:journals/ijon/LuoJZCLDL22,DBLP:conf/mm/MaXSYZJ22}.

\begin{figure}[t!]
\centering
\includegraphics[width=1\columnwidth]{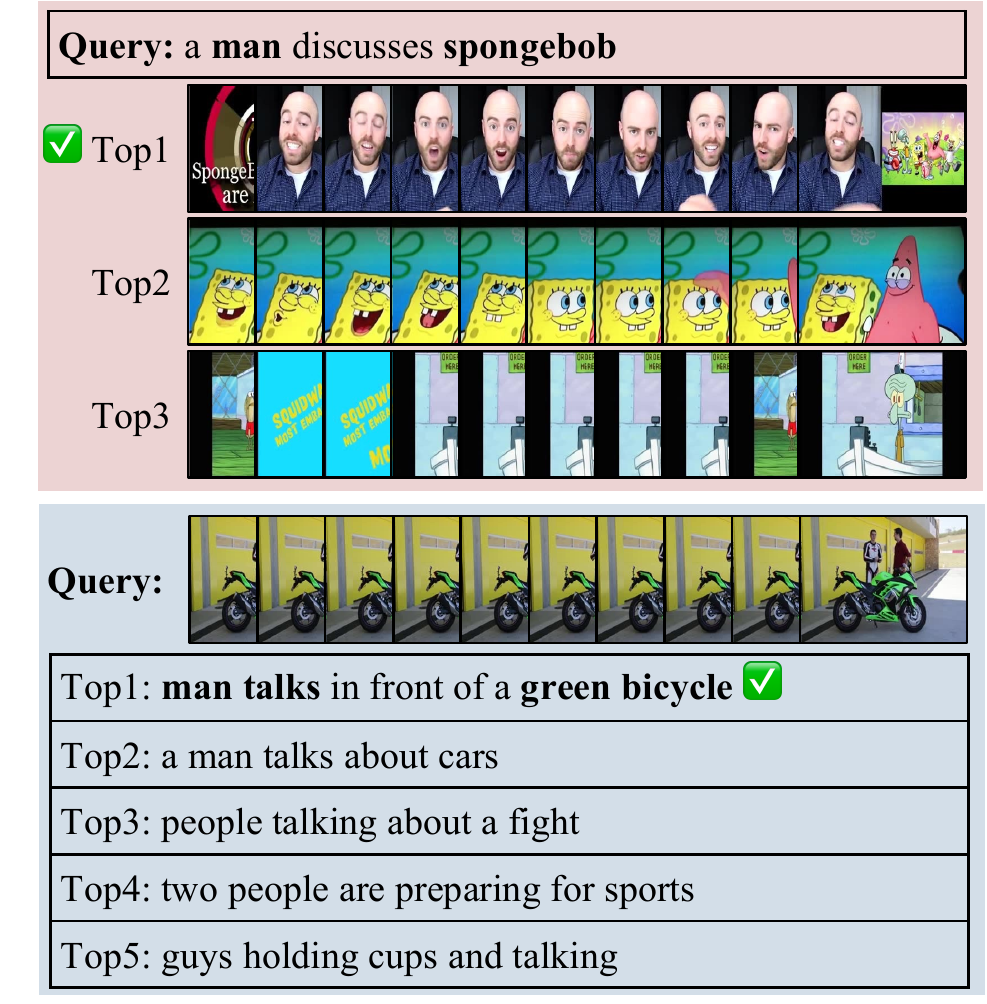}
\caption{Video-Text retrieval examples.}\label{fig: examples}
\end{figure}

\begin{figure}[t!]
\centering
\includegraphics[width=1\columnwidth]{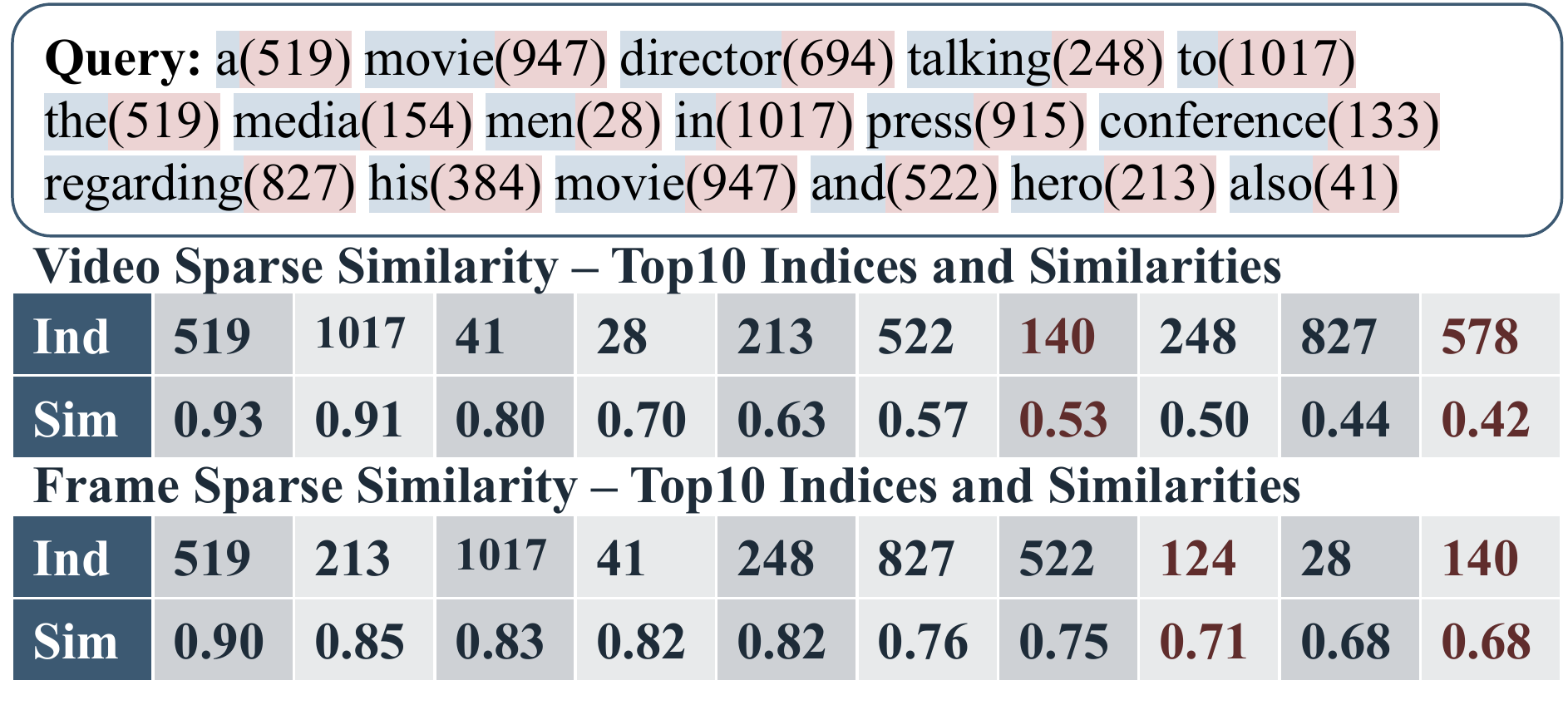}
\caption{An example of alignment on the sparse space. The index of the concepts is shown in the brackets.}\label{fig: examples sparse}
\end{figure}

\subsection{Qualitative Results}
To qualitatively validate the effectiveness of \ours\ and the alignment in the sparse space, we present examples of video-to-text and text-to-video retrieval on MSR-VTT in \Cref{fig: examples,fig: V2T,fig: T2V}, and the alignment in sparse space in \Cref{fig: examples sparse}, respectively. 
The retrieval results show the satisfactory performance of \ours, benefiting from multi-space multi-grained similarity. 
Notably, \ours\ demonstrates precise identification of the color (\textit{green}), objects (\textit{bicycle}), and humans (\textit{a man}), indicating its proficiency in capturing intricate details. 
In \Cref{fig: examples sparse}, we notice that, the video and frame features are perfectly aligned with the corresponding sparse concepts as exhibiting high similarities. 

\section{Conclusion}
In this paper, to better align video and text modalities, we proposed a multi-space, multi-grained video-text retrieval framework, \ours. 
Specifically, \ours\ aligned different modalities in a fine-grained shared sparse space, which is initialized with a finite number of concept clusters consisting of a number of basic concepts (words) and updated in a supervised fashion with the guide of text. 
Besides, \ours\ employed frame (fine-grained) and video (coarse-grained) features to encourage models to perform multi-grained similarity alignment. 
Finally, we conducted extensive experiments on three representative video-text retrieval benchmarks, showing the superiority of \ours. 

\section*{Limitations}
In the future, it would be promising to seek more fine-grained alignment, such as instance (object)-level or word-level alignment, for aligning different modalities. 
Moreover, our experiment focused solely on the application of sparse retrieval in video-text retrieval. 
It would be great to see whether sparse retrieval can help other cross-modal retrieval tasks, \eg, audio-text, image-text, audio-video, and audio-image retrieval. 
Additionally, incorporating more detailed information such as the relationship between different objects and frames would be beneficial for the video-text retrieval problem.

Regarding the sparse space, we notice that some sparse concepts are retrieved a lot during the training procedure which might lead to the emergence of hubness~\cite{DBLP:journals/jmlr/RadovanovicNI10}. 
Investigating improved clustering methods to mitigate hubness would be an interesting direction for future research. 
That might be due to the KNN clustering strategy and in the future and introducing better clustering strategies might be able to reduce the hubness issue, such as weighted KNN, semantic-based KNN, or part-of-speech tagging-based KNN.

\bibliography{anthology,custom,references}
\bibliographystyle{acl_natbib}
\clearpage

\appendix

\begin{table*}[h!]
\centering
\resizebox{\textwidth}{!}{%
\begin{tabular}{l|ccccc|ccccc}
\toprule
             & \multicolumn{5}{c}{Text-to-Video Retrieval}                                        & \multicolumn{5}{c}{Video-to-Text Retrieval}                                        \\
             & R@1$\uparrow$ & R@5$\uparrow$ & R@10$\uparrow$ & MdR$\downarrow$ & MnR$\downarrow$ & R@1$\uparrow$ & R@5$\uparrow$ & R@10$\uparrow$ & MdR$\downarrow$ & MnR$\downarrow$ \\
\midrule
\ours\ (ViT-B/32) w. multi-label classification                  &   47.0 & 73.6 & \textbf{82.9} & 2.0 & \textbf{12.5} & 45.5 & \textbf{73.8} & \textbf{82.8} & 2.0 & \textbf{8.7}   \\
\rowcolor{green!10}\ours\ (ViT-B/32) w. cosine     & \textbf{49.1}   & \textbf{73.9}   & 82.8  & 2.0    & 13.5    & \textbf{46.9}   & \textbf{73.8}   & 82.1   & 2.0    & 9.3             \\

\bottomrule
\end{tabular}%
}
\caption{Ablation study on the calculation of similarity between video and frame representations and cluster concepts. 
}
\label{tab: mls}
\end{table*}

\begin{table*}[h!]
\centering
\resizebox{\textwidth}{!}{%
\begin{tabular}{l|ccccc|ccccc}
\toprule
             & \multicolumn{5}{c}{Text-to-Video Retrieval}                                        & \multicolumn{5}{c}{Video-to-Text Retrieval}                                        \\
             & R@1$\uparrow$ & R@5$\uparrow$ & R@10$\uparrow$ & MdR$\downarrow$ & MnR$\downarrow$ & R@1$\uparrow$ & R@5$\uparrow$ & R@10$\uparrow$ & MdR$\downarrow$ & MnR$\downarrow$ \\
\midrule
\ours\ (ViT-B/32) w/o anchor                  &   47.8&72.9&82.3&2.0&\textbf{13.4}&46.4&\textbf{74.9}&\textbf{82.1}&2.0&\textbf{9.1}   \\
\rowcolor{green!10}\ours\ (ViT-B/32) w. anchor     & \textbf{49.1}   & \textbf{73.9}   & \textbf{82.8}  & 2.0    & 13.5    & \textbf{46.9}   & 73.8   & \textbf{82.1}   & 2.0    & 9.3             \\

\bottomrule
\end{tabular}%
}
\caption{Ablation study on the instruction of text, \ie, generating $\sentencecodebook$ using the similarity or the text. 
``w. anchor'' refers to obtain $\sentencecodebook$ by text as Eq.~\eqref{eq:sentencecodebook}.
``w/o anchor'' refers to obtain $\sentencecodebook$ by the similarity between sentence representations and concepts $C$ as Eq.~\eqref{eq: similarity sentence codebook}
}
\label{tab: wot gold}
\end{table*}

\begin{table*}[ht!]
\centering
\resizebox{\textwidth}{!}{%
\begin{tabular}{cc|ccccc|ccccc}
\toprule
          &         & \multicolumn{5}{c|}{Text-to-Video Retrieval}                                       & \multicolumn{5}{c}{Video-to-Text Retrieval}                                         \\
$\ell_{align}$ & $\ell_{alignsim}$ & R@1$\uparrow$ & R@5$\uparrow$ & R@10$\uparrow$ & MnR$\downarrow$ & MeanR$\downarrow$ & R@1$\uparrow$ & R@5$\uparrow$ & R@10$\uparrow$ & MnR$\downarrow$ & MeanR$\downarrow$ \\ \midrule
        &        & 48.0          & 72.9          & 82.4           & 2.0             & 13.5              & 45.4          & 73.2          & 82.1           & 2.0             & 9.3               \\
\Checkmark     &        & 48.0          & 73.5          & 82.7           & 2.0             & \textbf{13.4}              & \textbf{47.1}          & \textbf{74.2}          & \textbf{82.9}           & 2.0             & \textbf{9.1}               \\
        & \Checkmark    & 47.4          & 73.5          & 82.7           & 2.0             & 13.5              & 46.8          & 73.2          & 82.2           & 2.0             & 9.2               \\
\rowcolor{green!10}\Checkmark     & \Checkmark    & \textbf{49.1}          & \textbf{73.9}          & \textbf{82.8}           & 2.0             & 13.5              & 46.9          & 73.8          & 82.1           & 2.0             & 9.3               \\\bottomrule
\end{tabular}%
}
\caption{Ablation study of $\ell_{align}$ and $\ell_{alignsim}$ on MSR-VTT based on \ours~(ViT-B/32).}
\label{tab: two losses}
\end{table*}

\begin{table*}[ht!]
\centering
\resizebox{\textwidth}{!}{%
\begin{tabular}{cc|ccccc|ccccc}
\toprule
          &         & \multicolumn{5}{c|}{Text-to-Video Retrieval}                                       & \multicolumn{5}{c}{Video-to-Text Retrieval}                                         \\
$\alpha$ & $\beta$ & R@1$\uparrow$ & R@5$\uparrow$ & R@10$\uparrow$ & MnR$\downarrow$ & MeanR$\downarrow$ & R@1$\uparrow$ & R@5$\uparrow$ & R@10$\uparrow$ & MnR$\downarrow$ & MeanR$\downarrow$ \\ \midrule
\rowcolor{green!10}0.02     & 0.01    & \textbf{49.1}          & 73.9          & 82.8           & 2.0             & 13.5              & \textbf{46.9}          & 73.8          & 82.1           & 2.0             & 9.3               \\
\midrule
0.02     & 0.02    & 48.5          & 73.8          & \textbf{83.2}           & 2.0             & 14.0              & 46.3          & 73.1          & 82.1           & 2.0             & 9.4               \\
0.02     & 0.05    & 47.6          & 72.7          & 82.4           & 2.0             & 14.0              & 45.8          & \textbf{74.0}          & 82.2           & 2.0             & 9.2               \\
0.02     & 0.1     & 47.7          & 72.3          & 82.9           & 2.0             & 13.4              & 45.3          & 73.6          & \textbf{83.3}           & 2.0             & \textbf{9.0}               \\
\midrule
                          0.01     & 0.01    & 47.6          & 74.0          & 82.7           & 2.0             & 13.8              & 46.7         & 73.5    & 82.2                  & 2.0             & 9.5               \\
                                                    0.05     & 0.01    & 48.1          & 73.6          & 83.1           & 2.0             & \textbf{13.2}              & 46.3          & 72.9          & 82.7           & 2.0             & 9.1               \\
0.1      & 0.01    & 47.9          & \textbf{74.2}          & 82.3           & 2.0             & 13.3              & 46.3          & 73.4          & 82.5           & 2.0             & 9.1               \\
                        \bottomrule
\end{tabular}%
}
\caption{Ablation study of $\alpha$ and $\beta$ on MSR-VTT based on \ours~(ViT-B/32).}
\label{tab: two losses changing}
\end{table*}

\begin{table*}[ht!]
\centering
\resizebox{\textwidth}{!}{%
\begin{tabular}{cccc|cccc|ccccc|cccccc}
\toprule
 \multicolumn{4}{c|}{Dense Space} & \multicolumn{4}{c|}{Sparse Space} & \multicolumn{5}{c|}{Text-to-Video Retrieval}                                          & \multicolumn{5}{c}{Video-to-Text Retrieval}                                                                                     \\
                           S-V    & S-F  & W-V    & W-F   & S-V       & S-F     & W-V    & W-F   & R@1$\uparrow$ & R@5$\uparrow$ & R@10$\uparrow$ & MdR$\downarrow$ & MnR$\downarrow$ & R@1$\uparrow$ & R@5$\uparrow$ & R@10$\uparrow$ & MdR$\downarrow$ & MnR $\downarrow$ \\\midrule
\rowcolor{green!10}                           \Checkmark      & \Checkmark  &&   & \Checkmark         & \Checkmark  &&      &      \textbf{49.1}   & \textbf{73.9}   & \textbf{82.8}   & 2.0    & 13.5    & \textbf{46.9}   & {73.8}   & \textbf{82.1}   & 2.0    & \textbf{9.3}         \\
                  \Checkmark      & \Checkmark  &\Checkmark &  \Checkmark  & \Checkmark         & \Checkmark  &\Checkmark &  \Checkmark     &   48.3 & 73.8 & 82.7 & 2.0 & \textbf{13.0} & 46.6 & \textbf{74.1} & \textbf{82.1} & 2.0 & 9.4      \\\midrule
\rowcolor{gray!10} \multicolumn{8}{l}{{X-CLIP}}    & \multicolumn{1}{|c}{46.1}          & {74.3}          & {83.1}           & 2.0             & {13.2}            & {46.8}          & 73.3          & {84.0}           & 2.0             & {9.1}             \\

                          \bottomrule
\end{tabular}%
    }
\caption{Retrieval performance with different similarities on MSR-VTT using \ours\ with the base model of ViT-B/32. ``S-V'', ``S-F'', ``W-V'', and ``W-F'' represent Sentence-Video (coarse-grained), Sentence-Frame (fine-grained), Word-Video (fine-grained), and Word-Frame (fine-grained) similarities.}
\label{tab: with words}
\end{table*}

\begin{figure*}[t!]
\centering
\includegraphics[width=1\textwidth]{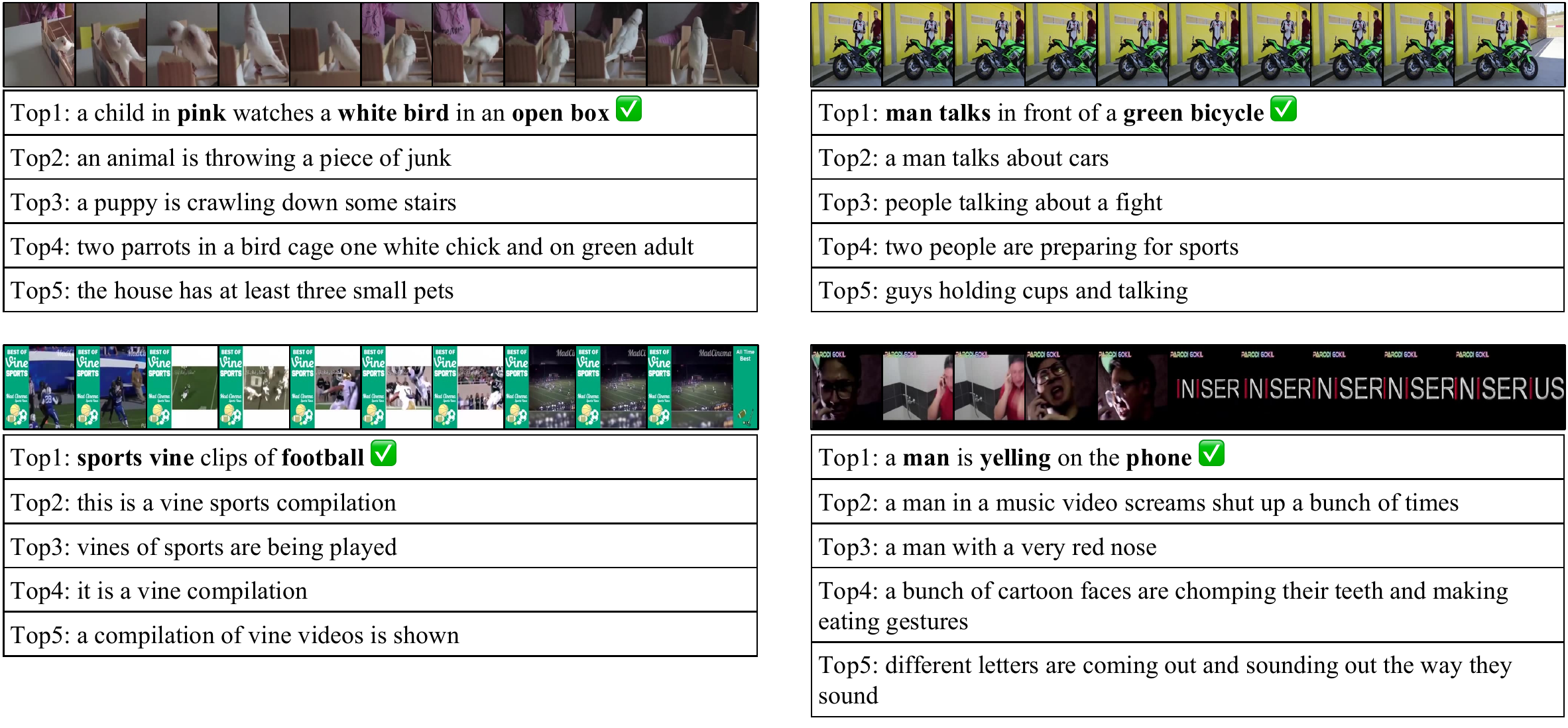}
\caption{Top-5 video-to-text retrieval results on MSR-VTT. 
}\label{fig: V2T}
\end{figure*}

\begin{figure*}[t!]
\centering
\includegraphics[width=1\textwidth]{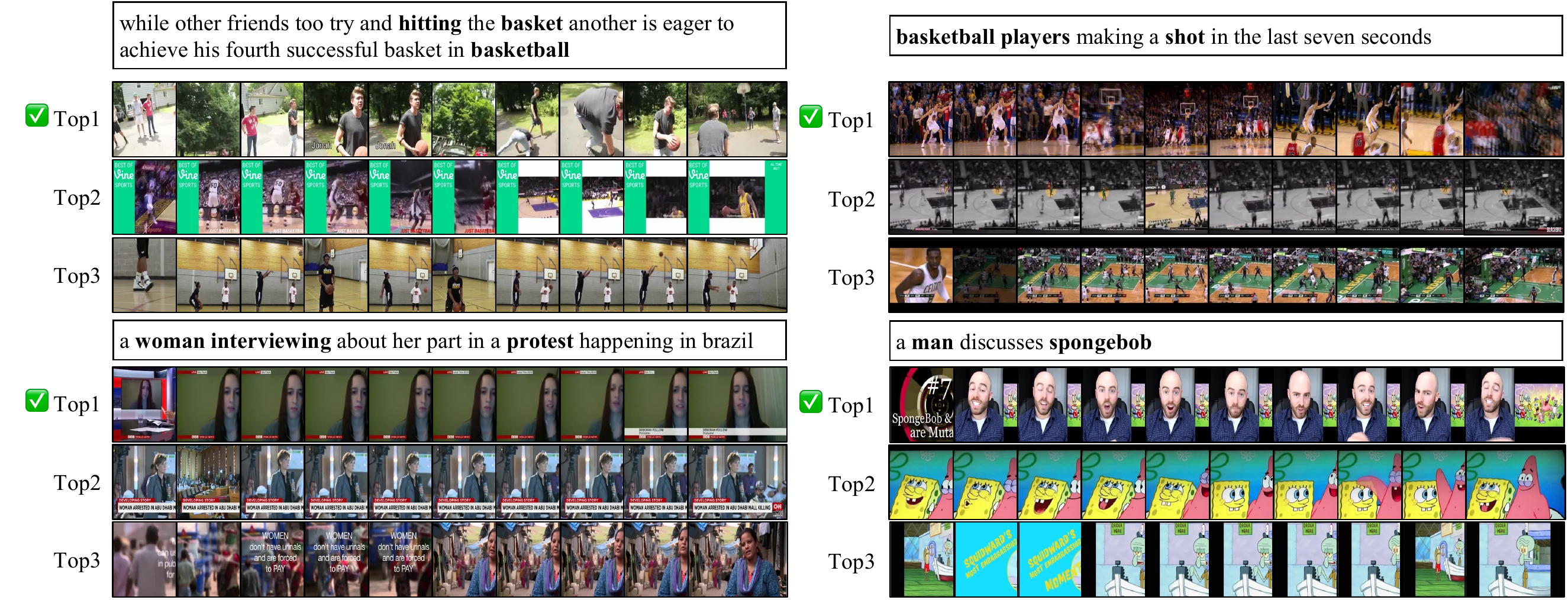}
\caption{Top-3 text-to-video retrieval results on MSR-VTT. 
}\label{fig: T2V}
\end{figure*}

\begin{figure*}[t!]
\centering
\includegraphics[width=1\textwidth]{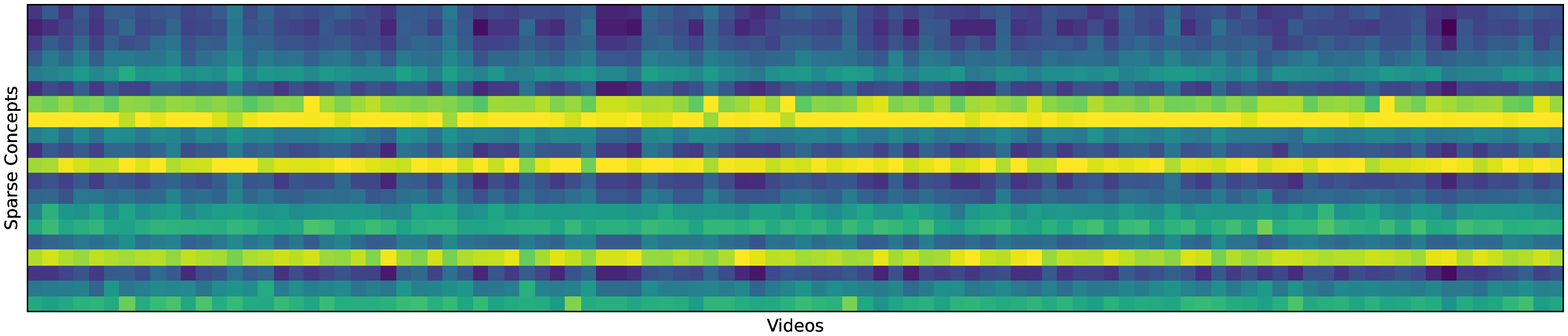}
\caption{The activation of 20 sparse concepts by 100 randomly selected videos. 
}\label{fig: concept activation}
\end{figure*}

\section{Experiments}

\subsection{Datasets Details}

\textbf{MSR-VTT}~\cite{DBLP:conf/cvpr/XuMYR16} contains 10,000 videos with length varying from 10 to 32
seconds, each paired with about 20 human-labeled captions. 
Following the evaluation protocol from previous works~\citet{DBLP:conf/eccv/YuKK18,DBLP:conf/iccv/MiechZATLS19}, we use the training-9k / test 1k-A splits for training and testing respectively. 

\textbf{MSVD}~\cite{chen-dolan-2011-collecting} contains 1,970 videos with a split of 1200, 100, and 670 as the train, validation, and test set, respectively.
The duration of videos varies from 1 to 62 seconds. 
Each video is paired with 40 English captions. 

\textbf{ActivityNet}~\cite{caba2015activitynet} is consisted of 20,000 Youtube videos with 100,000 densely annotated descriptions. 
For a fair comparison, following the previous setting~\cite{DBLP:journals/ijon/LuoJZCLDL22,DBLP:conf/eccv/Gabeur0AS20}, we concatenate all captions together as a paragraph to perform a video-paragraph retrieval task by concatenating all the descriptions of a video. 
Performances are reported on the ``val1'' split of the ActivityNet.

\subsection{Implementation Details and Evaluation Protocols}

Following \citet{DBLP:journals/ijon/LuoJZCLDL22,DBLP:conf/mm/MaXSYZJ22}, we use a standard vision transformer~\cite{DBLP:conf/iclr/DosovitskiyB0WZ21} with $12$ layers which are initialized with the public CLIP~\cite{DBLP:conf/icml/RadfordKHRGASAM21} checkpoints. 
We directly use the text encoder of CLIP as our text encoder which is also initialized with the public CLIP checkpoints.

We set the query, key, and value projection dimension size as $512$ to match CLIP’s output dimension and we initialize our logit scaling parameter $\lambda$ with the value from the pre-trained CLIP model. 
All models are optimized for 5 epochs on MSR-VTT and MSVD, and for ActivityNet, the models are trained for 20 epochs. We use AdamW~\cite{DBLP:conf/iclr/LoshchilovH19} with a weight decay of 0.2 and decay the learning rate using a cosine schedule~\cite{DBLP:conf/iclr/LoshchilovH17}, following the method used in CLIP~\cite{DBLP:conf/icml/RadfordKHRGASAM21}. 
For all experiments, we uniformly sample 12 frames from every video, resizing each frame to 224x224 as per previous works~\cite{DBLP:journals/ijon/LuoJZCLDL22,DBLP:conf/mm/MaXSYZJ22}.
we set $n_{codes}=1024$ following DiscreteCodebook~\cite{liu-etal-2022-cross}. 
To evaluate the retrieval performance of our proposed model, we use recall at Rank K (R@K, higher is better), median rank (MdR, lower is better), and mean rank (MnR, lower is better) as retrieval metrics, which are widely used in previous retrieval works~\cite{DBLP:conf/icml/RadfordKHRGASAM21,DBLP:journals/ijon/LuoJZCLDL22,DBLP:conf/mm/MaXSYZJ22}.

\subsection{Ablation Studies}

\textbf{Evaluating the calculation of similarity between video and frame representations and cluster concepts in \ours}. 
In \ours, we use cosine similarity to calculate $sim^{f}$ and $sim^{\video}$. 
Another way of calculating $sim^{f}$ and $sim^{\video}$ might be using multi-label classification. 
To compare the effect of multi-label classification and cosine similarity, we conduct experiments using two multi-layer perceptrons (MLPs) with two layers and the $ReLU$ activation to predict the similarity between video and frame representations and cluster concepts. 
Two MLPs are also trainable. 
Quantitative results are shown in Table~\ref{tab: mls}. 
Our quantitative results, shown in Table~\ref{tab: mls}, indicate that the use of MLPs decreases R@1 on text-to-video and video-to-text retrieval. 
This suggests that cosine similarity is more suitable for VTR.

\smallskip \noindent 
\textbf{Evaluating the importance of supervised alignment in \ours}. 
In \ours, the aligned sentence representation $\sentencecodebook$ is obtained from the text as in Eq.~\eqref{eq:sentencecodebook}. 
This process aligns the sentence representation based on the instruction of the text. 
By doing so, the aligned sentence representation $\sentencecodebook$ can serve as the supervision (an anchor) for aligning video and frame features, providing a reference point for the alignment of different modalities. 
To investigate the importance of placing an anchor $\sentencecodebook$ for better alignment, we compare it to obtaining aligned sentence representation through the similarity between concept clusters $C$ and sentence feature $\textbf{r}^{\text}$. 
This alternative approach allows us to evaluate the effectiveness of using an anchor for alignment and to understand how it improves the performance of the model. 
To investigate the alternative approach of obtaining aligned sentence representation without an anchor, we calculate the sentence sparse space similarity $sim^{\text} \in \mathbb{R}^{1\times n_{c}}$ by calculating the cosine similarity between sentence representations and concepts as $sim^{\text}_j = \operatorname{cos}(\sentencefeature, C_j)$, where $sim^{\text}_j$ is the $j$-th element of $sim^{\text}$, $C_j$ is the $j$-th row of $C$, and $\operatorname{cos}$ is the cosine similarity.
The aligned sentence representation $\textbf{r}^{\text}$ without the instruction of text is obtained by matrix multiplication as follows:
\begin{align}
\mathbf{r}^{\text} = sim^{\text} C / \|sim^{\text}\|_1 \label{eq: similarity sentence codebook},
\end{align}
where $sim^{\text}$ is the similarity between sentence representations and concepts.
The results of this comparison can be found in Table~\ref{tab: wot gold}. The experimental results show that with the ``anchor'', \ours\  can better align different modalities as R@1, R@5, and R@10 on text-to-video retrieval and R@1 on video-to-text retrieval have greatly improved, indicating that the supervised (anchor-based) alignment is crucial for better performance of the model.

\smallskip \noindent 
\textbf{Effect of losses and hyperparameter sensitivity}.
To further demonstrate the effectiveness of the two proposed losses designed for aligning different modalities in the shared sparse space, we conduct experiments to compare the performance of these losses. 
The quantitative results of these experiments are shown in Table~\ref{tab: two losses}. 
The results indicate that adding both losses simultaneously achieves the best performance on the MSR-VTT dataset. 
When using only one loss, the performance on text-to-video retrieval is comparable to the method without using both losses on text-to-video retrieval, but outperforms the method without the two losses on video-to-text retrieval. 
Specifically, when using two losses, R@1 on text-to-video retrieval and video-to-text retrieval is improved by 1.1 and 1.5, respectively. 
Additionally, all the other metrics, such as R@5 and R@10, are also improved, demonstrating the power of the two proposed losses in aligning different modalities in the shared sparse space.
To gain a better understanding of the sensitivity of \ours\ with respect to the two hyperparameters, $\alpha$ and $\beta$, we conduct a series of experiments with different settings of $\alpha$ and $\beta$ as shown in Table~\ref{tab: two losses}.
The results of these experiments demonstrate that, even with varying settings of $\alpha$ and $\beta$, the video-text retrieval performance remains consistent, indicating that the model is robust and not highly sensitive to these hyperparameters. 
This suggests that \ours\ is able to achieve good performance across a wide range of settings for these hyperparameters, making it easy to adjust and optimize for specific use cases. 
Additionally, this also suggests that \ours\ is not overly dependent on precise values of these hyperparameters, and is instead able to leverage the more important underlying features and patterns in the data.

\smallskip \noindent 
\textbf{Are word-level features necessary?}
To investigate the necessity of word-level features, we introduce word-level dense and sparse representations, along with word-frame and word-video similarities, into the dense and sparse spaces. The results are presented in \Cref{tab: with words}. Notably, we observe a decrease in performance when incorporating word-level contrast in both dense and sparse spaces, indicating possible feature redundancy. Moreover, our approach, which incorporates word-level contrast, can be viewed as an extension of X-CLIP~\cite{10.1145/3503161.3547910} with the shared sparse space. 
We notice that contrasting representations in the aligned sparse space enhances the retrieval performance of X-CLIP.

\subsection{Aligning Examples}

To show the effectiveness of \ours, we illustrate some examples of video-to-text and text-to-video retrieval examples in \Cref{fig: examples,fig: V2T,fig: T2V}. 
We notice that \ours\ is able to align some important concepts between video and text for precise retrieval. 
For example, in the bottom-left video-to-text result (Figure~\ref{fig: V2T}), the biggest difference between the top 5 retrieved texts is ``football''. 
By precisely capturing ``football'' in the video, \ours\ is able to give higher logits to the sentences that contain ``football''. 
Additionally, in the last (bottom-right) text to video result (Figure~\ref{fig: T2V}), we notice that, by understanding ``man'' and ``discuss'', \ours\ is able to distinguish the top 3 retrieved videos and select the one in which a man appears. 
This empirically shows that \ours\ performs well in visual and textual content understanding, benefiting from multi-space and multi-grained similarity.

Moreover, we visualize the activation of sparse concepts by videos in \Cref{fig: concept activation}. 
We notice that, some hub sparse concepts are frequently retrieved while some are not retrieved a lot, which might be due to the KNN clustering. 
Moreover, we notice that the difference between activations from videos are separable.

\end{document}